\documentclass[letterpaper]{article} 
\usepackage{aaai2026}  
\usepackage{times}  
\usepackage{helvet}  
\usepackage{courier}  
\usepackage[hyphens]{url}  
\usepackage{graphicx} 
\usepackage{natbib}  
\usepackage{caption} 
\usepackage{algorithm}
\usepackage{algorithmic}

\usepackage{newfloat}
\usepackage{listings}
\DeclareCaptionStyle{ruled}{labelfont=normalfont,labelsep=colon,strut=off} 
\floatstyle{ruled}
\newfloat{listing}{tb}{lst}{}
\floatname{listing}{Listing}
\title{Task-Specific Distance Correlation Matching for \\ Few-Shot Action Recognition}
\author{
    Fei Long\textsuperscript{\rm 1}\equalcontrib,
    Yao Zhang\textsuperscript{\rm 1}\equalcontrib,
    Jiaming Lv\textsuperscript{\rm 1},
    Jiangtao Xie\textsuperscript{\rm 1},
    Peihua Li\textsuperscript{\rm 1}\thanks{Corresponding Author.}
}
\affiliations{
    \textsuperscript{\rm 1}School of Information and Communication Engineering, Dalian University of Technology\\

    longfei121@mail.dlut.edu.cn,\; zyemail@mail.dlut.edu.cn,\; Peihua Li@dlut.edu.cn
}

\usepackage{bibentry}
\usepackage{xcolor}
\usepackage{enumitem}
\usepackage{subcaption}

\usepackage{listings}
\usepackage{booktabs}      
\usepackage{enumitem}
\usepackage{arydshln}
\usepackage{multirow}
\usepackage{amsmath}
\usepackage{amssymb}

\begin{document}
	
	\maketitle
	
	\begin{abstract}
		Few-shot action recognition (FSAR) has recently made notable progress through set matching and efficient adaptation of large-scale pre-trained models. However, two key limitations persist. First, existing set matching metrics typically rely on cosine similarity to measure inter-frame linear dependencies and then perform matching with only instance-level information, thus failing to capture more complex patterns such as nonlinear relationships and overlooking task-specific cues. Second, for efficient adaptation of CLIP to FSAR, recent work performing fine-tuning via skip-fusion layers (which we refer to as side layers) has significantly reduced memory cost. However, the newly introduced side layers are often difficult to optimize under limited data conditions. To address these limitations, we propose TS-FSAR, a framework comprising three components: (1) a visual Ladder Side Network (LSN) for efficient CLIP fine-tuning; (2) a metric called Task-Specific Distance Correlation Matching (TS-DCM), which uses $\alpha$-distance correlation to model both linear and nonlinear inter-frame dependencies and leverages a task prototype to enable task-specific matching; and (3) a Guiding LSN with Adapted CLIP (GLAC) module, which regularizes LSN using the adapted frozen CLIP to improve training for better $\alpha$-distance correlation estimation under limited supervision. Extensive experiments on five widely-used benchmarks demonstrate that our TS-FSAR yields superior performance compared to prior state-of-the-arts.
	\end{abstract}

	\section{Introduction}\label{sec:intro}

	Action recognition, as a fundamental task in visual learning, has attracted widespread attention and achieved remarkable success in recent years~\cite{tsn, i3d, nonlocal, trn, tsm, vivit, dual_path, riva}. However, training for this task typically requires large-scale labeled video datasets, which are costly to collect and often impractical in real-world scenarios. To overcome this limitation, few-shot action recognition (FSAR) seeks to recognize previously unseen classes from only limited labeled examples, and has thus emerged as an active and growing research area within the community~\cite{cmn++, otam, strm, trajectory, team}.

	In few-shot action recognition, most existing methods focus on designing more effective metrics and exploring efficient adaptation from large-scale pre-trained models—such as ImageNet-1K pretrained models~\cite{resnet, vit} or CLIP~\cite{clip}. To obtain a more discriminative metric, it is particularly important to consider how the distance between query and support prototypes is computed. In contrast to early methods~\cite{otam} that formulate the distance computation as a temporal alignment problem, several recent methods reformulate this as a matching problem between two sets of features, employing techniques such as Hausdorff distance~\cite{hyrsm} or optimal transport~\cite{mtfan, tsam} to compute the distance. Despite effectiveness, they still suffer from two primary limitations. First, they typically rely on cosine similarity to construct a distance matrix that captures inter-frame relationships between query and support. However, cosine similarity is approximately equivalent to Pearson Correlation Coefficient~\cite{correlation}, which can only capture linear relationships and thus fails to model more complex correlations such as nonlinear dependencies. Second, they perform matching based on instance-level information without considering task-specific cues, whose effectiveness has been demonstrated in prior works~\cite{mtfan, hyrsm, task_adapter}. This limits their ability to achieve more accurate matching. To enable efficient adaptation of CLIP in FSAR, recent works~\cite{ds2tadapter, mafsar, tsam} have introduced parameter-efficient adapters instead of full fine-tuning~\cite{clipfsar}. However, these approaches still require backpropagation through the backbone, leading to considerable GPU memory consumption. In contrast, EMP-Net~\cite{empnet} proposes to fine-tune the newly introduced skip-fusion layers that take as input the activation features from the frozen CLIP, thereby avoiding backbone backpropagation and reducing memory usage. This design shares a similar principle with Ladder Side-Tuning~\cite{sidetuning}. Despite being memory-efficient, optimizing skip-fusion layers under limited data remains challenging, especially on static datasets that depend heavily on pretrained knowledge.
	
	Inspired by the observations above, we propose TS-FSAR, which introduces a novel metric that matches two set of features in a task-specific manner, and fine-tunes the CLIP vision encoder via a ladder side network (LSN) guided by the frozen CLIP equipped with an adapter. In particular, the metric we propose can be decomposed into two components. First, to capture comprehensive inter-frame dependencies between query and support, we adopt $\alpha$-distance correlation~\cite{bdc}, an extension of distance correlation~\cite{bdc} known for capturing both linear and nonlinear relationships, to compute a inter-frame $\alpha$-correlation matrix for each query-support pair. Second, to achieve task-specific matching, we employs a learnable generator which takes a query-specific task prototype as input to produce an matching matrix, which encodes the relative importance of relationships between different frames of the query and support videos. Then, the final similarity score is obtained by taking the inner product of the inter-frame $\alpha$-correlation matrix and the matching matrix. To improve the training of the LSN under limited data for reliable $\alpha$-distance correlation estimation, we obtain a distribution by applying softmax over the $\alpha$-distance correlations between LSN features and different class prototypes, and align it with the adapted frozen CLIP’s output distribution. To validate the effectiveness of our proposed TS-FSAR, we evaluated it on five standard datasets. The experimental results demonstrate that our method achieves superior performance compared to prior methods. Our contributions can be summarized as follows:
	
	\begin{itemize}
		\item We propose a novel metric, termed Task-specific Distance Correlation Matching (TS-DCM), for few-shot action recognition. It leverages $\alpha$-distance correlations to measure inter-frame relationships and employs a query-specific task prototype to perform task-specific matching.
		
		\item We introduce a Guiding LSN with Adapted CLIP (GLAC) module to guide the LSN using the output distribution of the adapted frozen CLIP, which helps improve the training of the LSN under limited data conditions, thereby contributing to more reliable $\alpha$-distance correlation estimation based on its features. 
		
		\item Our proposed TS-FSAR achieves leading performance across several few-shot action recognition benchmarks, with particularly significant improvements on temporally challenging datasets such as SSv2-Full.
	\end{itemize}
	
	\section{Related Work}

	\textbf{Few-shot Action Recognition~}
	In few-shot action recognition, a series of methods focus on enhancing video representations to improve generalization. AMeFu-Net~\cite{depth} leverages depth information through an adaptive normalization module with temporal asynchronization to enrich cross-modal representations. TRX~\cite{trx} builds query-specific class prototypes using cross-attention across ordered video sub-sequences. TA$^2$N~\cite{ta2n} introduces a two-stage alignment framework to mitigate temporal and spatial misalignment via temporal transformation and action coordination. MoLo~\cite{molo} proposes motion-augmented long-short contrastive learning to jointly capture long-range temporal dependencies and motion cues. Beyond these representation-oriented works, another branch of research aims to design more discriminative metrics. OTAM~\cite{otam} employs a variant of Dynamic Time Warping to align query–support sequences. HyRSM~\cite{hyrsm} formulates sequence distance computation as a set-matching problem and introduces bidirectional Mean Hausdorff matching. Following this paradigm, MTFAN~\cite{mtfan} and TSAM~\cite{tsam} compute the distance between query and support features using Optimal Transport. Our proposed metric differs from existing methods in two key aspects. First, instead of using cosine similarity to measure inter-frame relations, we adopt $\alpha$-distance correlation to achieve more comprehensive modeling of inter-frame dependencies. Second, unlike these metrics that follow a task-agnostic paradigm, our proposed TS-DCM perform matching between query and support videos in a task-specific manner, resulting in more accurate matching.

    \noindent \textbf{Side-Tuning for Video Understanding~}  
   Recently, efficiently adapting CLIP to domain-specific tasks has gained growing attention. Among existing methods, side-tuning~\cite{sidetuning,lst} provides a memory-efficient solution by attaching a lightweight side network to the frozen backbone, and has been widely adopted in recent studies, including those on video understanding. EVL~\cite{evl} introduces an efficient video recognition framework using frozen CLIP features, where a lightweight Transformer decoder and local temporal modules are used to capture spatial and temporal cues; STAN~\cite{stan} introduces a spatial-temporal auxiliary network with a branched architecture, which incorporates decomposed spatial-temporal modules to effectively contextualize multilevel CLIP features for video tasks; EMP-Net~\cite{empnet} leverages skip-fusion layers to integrate multi-stage CLIP features and performs multi-level post-reasoning in few-shot action recognition. Unlike these methods that focus on designing different side network architectures, we directly adopt a simple Ladder Side Network~\cite{lst} (LSN) for finetuning. During training, the output distribution of adapted frozen CLIP is used to guide the LSN, enabling more effective optimization under limited data conditions.

	\section{Methodology}
	
	\begin{figure*}[htbp]
		\centering	
		\begin{subfigure}[b]{0.90\linewidth}
			\centering
			\includegraphics[width=0.9\textwidth]{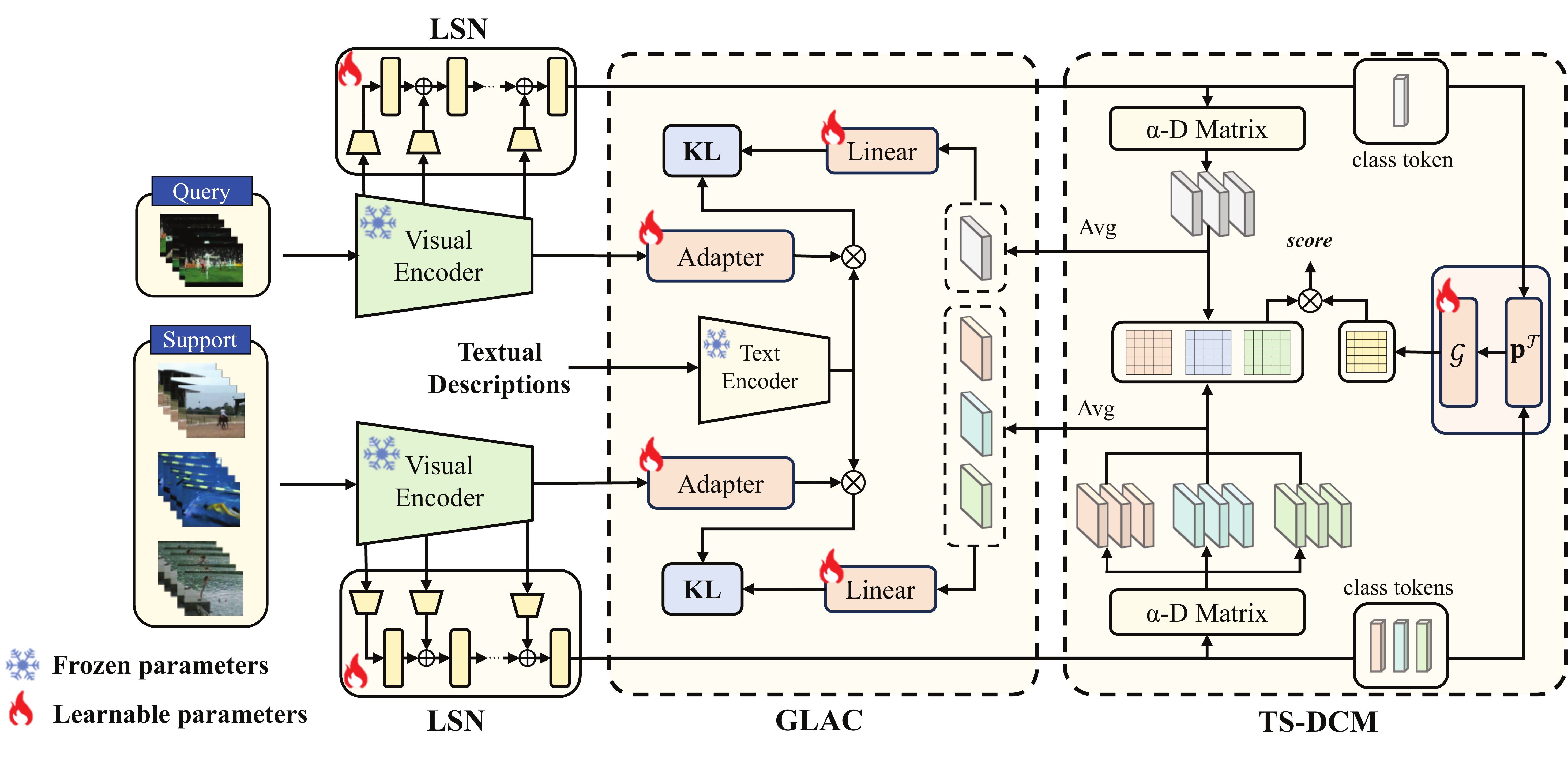}
		\end{subfigure}		
		\caption{Illustration of the proposed TS-FSAR framework, which comprises three components:
			(1) a Ladder Side Network (LSN) for memory-efficient fine-tuning of the CLIP visual encoder,
			(2) a metric named Task-Specific Distance Correlation Matching (TS-DCM) that leverages $\alpha$-distance correlation and a task prototype for more accurate query-support matching, and
			(3) a Guiding LSN with Adapted CLIP (GLAC) module that enhances LSN training under limited data to enable more reliable distance correlation estimation.} 
		\label{figure: overview}
	\end{figure*}

	\subsection{Problem Formulation}
	Few-shot action recognition (FSAR) aims to learn a model that can generalize to unseen action categories using only a few labeled examples. A FSAR dataset is typically divided into three disjoint subsets: $\mathcal{D}_{\text{train}}$, $\mathcal{D}_{\text{val}}$, and $\mathcal{D}_{\text{test}}$, each containing non-overlapping classes. To align with the evaluation scenario, FSAR models are typically trained in an episodic manner. Specifically, each episode comprises a support set $\mathcal{S} = \{(x_i, y_i)\}_{i=1}^{N_{\mathcal{S}}}$ and a query set $\mathcal{Q} = \{(x_i, y_i)\}_{i=1}^{N_{\mathcal{Q}}}$. The support set consists of $K$ labeled videos per class from $N$ classes, forming an $N$-way $K$-shot task with $N_{\mathcal{S}} = N \times K$ samples in total. The query set, denoted by $\mathcal{Q}$, includes $N_{\mathcal{Q}}$ samples to be classified. 
	
	\subsection{Overview of TS-FSAR}
	As illustrated in Figure~\ref{figure: overview}, the proposed TS-FSAR framework comprises three main components: a Ladder Side Network (LSN), a metric called Task-Specific Distance Correlation Matching (TS-DCM), and a Guiding LSN with Adapted CLIP (GLAC) module. In our framework, both query and support videos are processed in parallel, sharing all learnable parameters across the LSN, adapter, and linear layer.

	Take 1-shot for an example. Given a task, both query and support videos are first fed into the LSN to extract output features. Then, these features are processed by the TS-DCM to perform task-specific matching. It begins by computing the frame-level $\alpha$-distance ($\alpha$-D) matrices for the query and support videos independently, followed by deriving an inter-frame $\alpha$-distance correlation matrix for each query-support pair, capturing both linear and nonlinear dependencies between their frames. Then, a query-specific prototype is constructed by the class tokens of the support and query videos. This prototype is then passed into a learnable generator to produce a matching matrix that encodes the relative importance of inter-frame relationships between the query and support. Finally, the similarity scores for the query is obtained by computing the inner product between the matching matrix and the inter-frame $\alpha$-distance correlation matrices. 
	
	To improve the training of LSN under limited data for more reliable $\alpha$-distance correlation estimation, we introduce the GLAC module. First, both query and support videos are processed by frozen CLIP visual encoder, followed by an adapter that helps adapt CLIP to the video domain. The adapted visual features are then aligned with text embeddings to obtain a guidance distribution. Meanwhile, we average the frame-level $\alpha$-D matrices for each video to obtain video-level representation, and then obtain a distribution by applying softmax to the $\alpha$-distance correlations between the representation and learnable class prototypes implemented via a linear layer. The training of LSN is then guided by minimizing the KL divergence between these two distributions.

	\subsection{Ladder Side Network}

	Ladder Side Tuning (LST)~\cite{lst} is a memory-efficient fine-tuning strategy that introduces a lightweight and separate Ladder Side Network (LSN) alongside the backbone. The LSN receives dimension-reduced hidden features from the backbone via shortcut connections as input. To enable efficient adaptation, we employ the LSN to fine-tune the vision encoder of CLIP.

	During training, for each video, we align the visual outputs of the LSN with the corresponding text embeddings. Given a video with $T$ frames, the visual token embeddings of the $t$-th frame produced by the LSN can be denoted as $\mathbf{V}^{t} = [\mathbf{v}^{t}_{0}, \cdots, \mathbf{v}^{t}_{P}] \in \mathbb{R}^{(P+1) \times d}$, where $P$ is the number of patch tokens, and $d$ denotes the feature dimension. Next, we project the class token $\mathbf{v}^{t}_{0}$ through a linear layer to match the dimension of the text embeddings, yielding $\hat{\mathbf{v}}^{t}_{0}\in \mathbb{R}^{d'}$. We then average the class tokens $\hat{\mathbf{v}}^{t}_{0}$ from all frames to obtain the video-level representation $\mathbf{\widetilde{v}}$. Let $\mathbf{W} = [\mathbf{w}_1, \dots, \mathbf{w}_C]$ denote the text embeddings obtained from text encoder, where $\mathbf{w}_i \in \mathbb{R}^{d'}$ represents the embedding of the $i$-th class description, $C$ denotes the number of classes in $\mathcal{D}_{\text{train}}$. We then compute the cosine similarity between $\mathbf{\widetilde{v}}$ and $\mathbf{w}_i$, followed by a softmax operation to obtain the prediction. The resulting prediction is then used to define the training loss via cross-entropy, denoted as $\mathcal{L}_{\text{LSN}}$. \textit{More implementation details of the LSN  can be found in Appendix Sec.A.}

	\subsection{Task-Specific Distance Correlation Matching}

	Our proposed TS-DCM can be decomposed into two components: Inter-Frame $\alpha$-Distance Correlation (IF-D$^{\alpha}$C) and Task-Specific Matching (TSM). IF-D$^{\alpha}$C computes inter-frame correlation matrices between support and query videos using $\alpha$-distance correlation, capturing more comprehensive inter-frame dependencies. Then, TSM takes the task prototype as input to a learnable generator, which produces a matching matrix for performing task-specific matching between the query and support.
	
	\subsubsection{$\alpha$-Distance Correlation}
	$\alpha$-distance correlation is an extension of standard distance correlation~\cite{bdc}, which is known for effectively modeling both linear and nonlinear relationships between random variables. The $\alpha$-distance covariance between two random variables $\mathbf{X}$ and $\mathbf{Y}$ is defined as $\alpha$-weighted $L^2$ distance between their joint characteristic function $\varphi_{\mathbf{X}, \mathbf{Y}}(t, s)$ and the product of their marginal characteristic functions $\varphi_{\mathbf{X}}(t)$, $\varphi_{\mathbf{Y}}(s)$:
	\begin{equation}
		\text{DCov}^{2(\alpha)}(\mathbf{X}, \mathbf{Y}) 
		= \left\| \varphi_{\mathbf{X},\mathbf{Y}}(t,s) - \varphi_\mathbf{X}(t) \varphi_\mathbf{Y}(s) \right\|^2_{\alpha}
	\end{equation}
	The exponent parameter $\alpha \in (0, 2)$ serves to modulate the sensitivity to dependencies of varying scales. The $\alpha$-distance correlation is defined as the normalized version of $\alpha$-distance covariance, and serves to quantify the degree of dependence between two random variables.
	
	In the discrete case, one can follow the procedure in~\cite{bdc} to compute the empirical $\alpha$-distance correlation. Given two random variables $\mathbf{X}$ and $\mathbf{Y}$ with $m$ independent and identically distributed observations $\{(\mathbf{x}_1, \mathbf{y}_1), \ldots, (\mathbf{x}_m, \mathbf{y}_m)\}$, we first compute the $\alpha$-th power of pairwise Euclidean distances to obtain the matrices $\widehat{\mathbf{A}} = (\widehat{a}_{kl})$ and $\widehat{\mathbf{B}} = (\widehat{b}_{kl})$, as follows:
	\begin{equation} \label{eq: alpha}
		\widehat{a}_{kl} = \|\mathbf{x}_k - \mathbf{x}_l\|^{\alpha},  \widehat{b}_{kl} = \|\mathbf{y}_k - \mathbf{y}_l\|^{\alpha}  
	\end{equation}
	By applying standard double centering to matrix $\widehat{\mathbf{A}}$ and $\widehat{\mathbf{B}}$, we obtain the centered $\alpha$-distance matrices (referred to as $\alpha$-D matrix) for $\mathbf{X}$ and $\mathbf{Y}$, denoted as $\mathbf{A}$ and $\mathbf{B}$, respectively. Next, the $\alpha$-distance covariance  between $\mathbf{X}$ and $\mathbf{Y}$ can be obtained as:
	\begin{equation}
		\mathrm{DCov}^{2(\alpha)}(\mathbf{X}, \mathbf{Y}) = \frac{1}{m^{2}}\mathtt{tr}(\mathbf{A}\mathbf{B})
	\end{equation}
	Then, the $\alpha$-distance correlation is defined as:
	\begin{equation} \label{eq: DC}
		\mathrm{DCorr}^{2(\alpha)}(\mathbf{X}, \mathbf{Y}) = \frac{\mathtt{tr}(\mathbf{A}\mathbf{B})}{\sqrt{\mathtt{tr}(\mathbf{A}\mathbf{A})}\sqrt{\mathtt{tr}(\mathbf{B}\mathbf{B})}}
	\end{equation}
	
	\subsubsection{Inter-Frame $\alpha$-Distance Correlation}
	Let $\mathbf{V}^{i}_{\mathcal{S}} \in \mathbb{R}^{(P+1) \times d}$ and $\mathbf{V}^{j}_{\mathcal{Q}} \in \mathbb{R}^{(P+1) \times d}$ denote the features of the $i$-th frame from a support video and the $j$-th frame from a query video, respectively. By treating each column of $\mathbf{V}^{i}_{\mathcal{S}}$ and $\mathbf{V}^{j}_{\mathcal{Q}}$ as an observation of random vectors $\mathbf{X}$ and $\mathbf{Y}$, we compute the corresponding $\alpha$-D matrices, denoted as $\mathbf{A}^{i}$ and $\mathbf{B}^{j}$. Then, the $\alpha$-Distance Correlation $m_{ij}$ between the $i$-th frame of the support video and the $j$-th frame of the query video can be computed using Eq.~(\ref{eq: DC}). Based on this, the inter-frame $\alpha$-Distance Correlation matrix is formed as $\mathbf{M}^{\text{IF-}\mathrm{D^{\alpha}C}} = (m_{ij}) \in \mathbb{R}^{T \times T}$.

	\subsubsection{Task-Specific Matching}

	The matrix $\mathbf{M}^{\text{IF-}\mathrm{D}^{\alpha}\mathrm{C}}$ encodes inter-frame correlations between the support and query videos. The key to leveraging these correlations lies in designing an appropriate matching matrix that captures the relative importance between query and support frames. To achieve this, a task prototype is fed into a learnable generator to produce the matching matrix $\mathbf{M}^{\text{task}}$, enabling flexible and task-specific matching.

	Let $\mathbf{\widetilde{v}}_i^{\mathcal{S}}$ and $\mathbf{\widetilde{v}}^{\mathcal{Q}}$ denote the frame-wise averaged class token produced by LSN for the support video $x_i \in \mathcal{S}$ and a query video $x \in \mathcal{Q}$, respectively. Then we design a query-specific task prototype , which can be computed as :
	\begin{align}
		\mathbf{p}^{\mathcal{T}} =  \mathbf{\widetilde{v}}^{\mathcal{Q}} + \frac{1}{N_\mathcal{S}} \sum_{x_{i} \in \mathcal{S}} \mathbf{\widetilde{v}}_{i}^{\mathcal{S}} 
	\end{align}
	where $N_\mathcal{S}$ is the number of videos in the entire support set. Besides average-based fusion, we also explore alternative strategies such as concatenation and cross-attention, which will be discussed in Sec.~\ref{Ablation}.

	After getting the task prototype, we employ a learnable linear layer as the generator $\mathcal{G}(\cdot)$ to produce the task-specific matching matrix $\mathbf{M}^{\text{task}} \in \mathbb{R}^{T \times T}$:
	\begin{align}
		\mathbf{M}^{\text{task}} = \mathcal{G}(\mathbf{p}^{\mathcal{T}}) 
	\end{align}
	
	And then, the similarity score between the query video and the support video can be obtained as:
	\begin{align}
		score = \langle \mathbf{M}^{\text{task}}, \mathbf{M}^{\text{IF-}\mathrm{D^{\alpha}C}} \rangle 
	\end{align}
	where $\langle \cdot \rangle$ donote the inner product. 
	
	For a given query video, we compute the scores by matching it with the support video of each class. Applying softmax to these scores yields a prediction probability vector $\mathbf{s} = [s_1, s_2, \ldots, s_{N}]$. We then optimize the model by computing the cross-entropy loss between the ground-truth label and the prediction vector $\mathbf{s}$, denoted as $\mathcal{L}_{\text{TS-DCM}}$.

	\subsection{Guiding LSN with Adapted CLIP}

	Although finetuning with LSN is memory-efficient, optimizing a number of newly introduced layers with limited samples remains highly challenging. This directly influences the estimation of inter-frame $\alpha$-distance correlation derived from the output features of LSN. Motivated by this, we propose to guide the training of the LSN through alignment with the output distribution of the adapted frozen CLIP.
	

	For a given video, we average the frame-wise $\alpha$-D matrices to obtain the video-level representation $\widetilde{\mathbf{A}}_{\alpha-\mathrm{D}} \in \mathbb{R}^{d \times d}$. Then, to better adapt the estimation of the $\alpha$-distance correlation in TS-DCM, we initialize learnable weight matrix $\widetilde{\mathbf{W}}_i$ for each class $i$ as its $\alpha$-D matrix prototype. These weight matrices can be implemented using a linear layer. Then, the prediction is computed by taking the inner product between $\widetilde{\mathbf{A}}_{\alpha\text{-}\mathrm{D}}$ and $\widetilde{\mathbf{W}}_i$, followed by a softmax operation to produce the predicted probability vector $\mathbf{p} = [p_1, p_2, ..., p_{C}]$.
	
	To improve the guidance, we employ an adapter composed of a standard multi-head self-attention (MHSA) to help CLIP adapt to the video domain, where frame-level CLS tokens are fed into the adapter to model inter-frame dependencies. Subsequently, the class tokens from different frames produced by the adapter are averaged to form a video-level representation, denoted by $\widetilde{e}$. Then, we calculate the cosine similarity between $\widetilde{e}$  and the textual features output by text encoder, and apply a softmax function to obtain the guidance vector $\mathbf{q}=[q_1, q_2, ..., q_{C}]$. Then, we employ the Kullback–Leibler (KL) divergence to guide the learning process, and the loss is defined as follows:
	
	\begin{align}
		\mathcal{L}_{\text{GLAC-KL}} = \mathrm{KL}(\mathbf{p}\parallel \mathbf{q}) = \sum_{i=1}^{C} p_i \log \frac{p_i}{q_i}
	\end{align}

	Furthermore, ground-truth labels are employed to supervise both the CLIP and LSN branches through cross-entropy loss, thereby improving their individual training processes.
	
	\begin{align}
		\mathcal{L}_{\text{GLAC-CE}} =  -\sum_{i=1}^{C} y_i \log(p_i) +(-\sum_{i=1}^{C} y_i \log(q_i))
	\end{align}

	Finally, the total loss for GLAC module can be defined as:
	
	\begin{align}
		\mathcal{L}_{\text{GLAC}} = \mathcal{L}_{\text{GLAC-KL}} + \mathcal{L}_{\text{GLAC-CE}}
	\end{align}

	\subsection{Training Loss}

	The training loss of our TS-FSAR is composed of the three components described above: the vision-language alignment loss for the LSN, the TS-DCM loss, and the GLAC loss. Accordingly, the total loss can be formulated as:
	
	\begin{align}
		\mathcal{L} = \mathcal{L}_{\text{LSN}} + \lambda_1\mathcal{L}_{\text{TS-DCM}} + \lambda_2 \mathcal{L}_{\text{GLAC}}
	\end{align}
	where $\lambda_1$ and $\lambda_2$ are weights tuned on the validation set.
	
	\begin{table*}[h]
		\centering
		\setlength{\tabcolsep}{0.5mm}
		\renewcommand{\arraystretch}{1.1}
		\begin{tabular}{cccccccccccc}
			\hline
			& & \multicolumn{2}{c}{\textbf{SSv2-Full}} &
			\multicolumn{2}{c}{\textbf{SSv2-Small}} &
			\multicolumn{2}{c}{\textbf{HMDB51}} &
			\multicolumn{2}{c}{\textbf{UCF101}} &
			\multicolumn{2}{c}{\textbf{Kinetics}} \\
			\cline{3-12}
			\multirow{-2}{*}{\textbf{Method}} &
			\multirow{-2}{*}{\textbf{Backbone}} &
			\textbf{1-shot} & \textbf{5-shot} &
			\textbf{1-shot} & \textbf{5-shot} &
			\textbf{1-shot} & \textbf{5-shot} &
			\textbf{1-shot} & \textbf{5-shot} &
			\textbf{1-shot} & \textbf{5-shot} \\
			\hline
			OTAM~\cite{otam}         & IN-RN50         & 42.8 & 52.3 & 36.4 & 48.0 & 54.5 & 68.0 & 79.9 & 88.9 & 73.0 & 85.8 \\
			TRX~\cite{trx}           & IN-RN50         & 42.0 & 64.6 & 36.0 & 56.7 & 54.9 & 75.6 & 81.0 & 96.1 & 65.1 & 85.9 \\
			STRM~\cite{strm}         & IN-RN50         & 43.1 & 68.1 & 37.1 & 55.3 & 57.6 & 77.3 & 82.7 & 96.9 & 65.1 & 86.7 \\
			HyRSM~\cite{hyrsm}       & IN-RN50         & 54.3 & 69.0 & 40.6 & 56.1 & 60.3 & 76.0 & 83.9 & 94.7 & 73.7 & 86.1 \\
			HCL~\cite{hcl}           & IN-RN50         & 47.3 & 64.9 & 38.7 & 55.4 & 59.1 & 76.3 & 82.6 & 94.5 & 73.7 & 85.8 \\
			Nguyen~\cite{nguyen}     & IN-RN50         & 43.8 & 61.1 & –    & –    & 59.6 & 76.9 & 84.9 & 95.9 & 74.3 & 87.4 \\
			SloshNet~\cite{sloshnet} & IN-RN50         & 46.5 & 68.3 & –    & –    & 59.4 & 77.5 & 86.0 & 97.1 & 70.4 & 87.0 \\
			GgHM~\cite{gghm}         & IN-RN50         & 54.5 & 69.2 & –    & –    & 61.2 & 76.9 & 85.2 & 96.3 & 74.9 & 87.4 \\
			TEAM~\cite{team}         & IN-RN50         & – & – & – & – & 62.8 & 78.4 & 87.2 & 96.2 & 75.1 & 88.2 \\
			CLIP-FSAR~\cite{clipfsar}& CLIP-ViT-B/16   & 62.1 & 72.1 & 54.6 & 61.8 & 77.1 & 87.7 & 97.0 & 99.1 & 94.8 & 95.4 \\
			EMP-Net~\cite{empnet} & CLIP-ViT-B/16   & 63.1 & 73.0 & 57.1 & 65.7 & 76.8 & 85.8 & 94.3 & 98.2 & 89.1 & 93.5 \\
			MVP-shot~\cite{mvp} & CLIP-ViT-B/16   & - & - & 55.4 & 62.0 & 77.0 & 88.1 & 96.8 & 99.0 & 91.0 & 95.1 \\
			MA-FSAR~\cite{mafsar}    & CLIP-ViT-B/16   & 63.3 & 72.3 & 59.1 & 64.5 & 83.4 & 87.9 & 97.2 & 99.2 & 95.7 & 96.0 \\
			D$^{2}$ST-Adapter~\cite{ds2tadapter} & CLIP-ViT-B/16 & 66.7 & 81.9 & 55.0 & 69.3 & 77.1 & 88.2 & 96.4 & 99.1 & 89.3 & 95.5 \\
			TSAM~\cite{tsam} & CLIP-ViT-B/16   & 65.8 & 74.6 & \textbf{60.5} & 66.7 & 84.5 & \textbf{88.9} & 98.3 & 99.3 & 96.2 & 97.1 \\
			\hline
			TS-FSAR (Ours) & CLIP-ViT-B/16   & \textbf{75.1} & \textbf{83.5} & \textbf{60.5} & \textbf{70.3} & \textbf{85.0} & \textbf{88.9} &\textbf{98.7}  &\textbf{99.3}  & \textbf{96.3} &96.6  \\
			\hline
		\end{tabular}
		\caption{Performance comparison with state-of-the-art methods on standard benchmarks. IN denotes ImageNet.}
		\label{tab:comparsion with sotas}
	\end{table*}
	
	\section{Experiments}
	
	\subsection{Datasets}
	
	We evaluate our method on five commonly used benchmarks: SSv2-Full~\cite{ssv2}, SSv2-Small~\cite{ssv2}, Kinetics-100~\cite{k400}, UCF101~\cite{ucf101}, and HMDB51~\cite{hmdb51}. \textit{Please refer to Appendix Sec.B.1 for details about the dataset.}

	\subsection{Implementation}
	
	Following previous works~\cite{clipfsar, empnet, tsam}, we implement our framework using CLIP ViT-B/16 as the visual backbone and uniformly sample 8 frames to construct the input sequence for each video. The dimension of the LSN is set to 256, and the number of layers is aligned with the visual encoder, i.e., 12. Following TSAM~\cite{tsam}, we employ large language model to generate class-specific descriptions. During training, we use the AdamW optimizer with a weight decay of 0.1 and adopt a cosine learning rate schedule. The learning rate is initialized to 2e-4 for SSv2-Full, SSv2-Small, and HMDB51, and to 1e-4 in the 5-shot and 2e-4 in the 1-shot for Kinetics-100 and UCF101. We sample 50,000 training episodes for SSv2-Full, and 10,000 episodes for all other datasets. We report the average accuracy over 10,000 episodes. \textit{More implementation details are provided in Appendix Sec.B.2.} 

	\subsection{Comparison with State-of-the-Art Methods}
	
	We compare TS-FSAR with recent state-of-the-art approaches across both temporally-dependent and spatially-dependent FSAR benchmarks, as summarized in Table~\ref{tab:comparsion with sotas}.
	
	\textbf{Temporally-dependent datasets} On SSv2-Small, TS-FSAR performs comparably to the state-of-the-art method TSAM in 1-shot and outperforms existing methods by 1\% in 5-shot. On SSv2-Full, TS-FSAR significantly outperforms prior arts by 8.4\% in 1-shot and 1.6\% in 5-shot, demonstrating its superior ability to model complex temporal dependencies in few-shot scenarios.
	
	\textbf{Spatially-dependent datasets} On HMDB51, UCF101, and Kinetics-100, our TS-FSAR performs slightly better or comparably to the previous leading method TSAM, except for a minor drop in 5-shot on Kinetics-100. We suspect that this is due to the relatively weak temporal dynamics of these datasets. Compared to SSv2-Full and SSv2-Small, these datasets are more aligned with CLIP’s pretraining distribution, which consists of static images lacking temporal structure, and thus rely more heavily on CLIP’s pretrained weights. As a result, fine-tuning the LSN on their limited training set becomes more challenging. Compared to EMP-Net, which also employs a side network, our method benefits from the GLAC module, achieving 1.1\%$\sim$8.2\% performance gains. However, this design still cannot fully resolve the optimization challenge under such conditions.
	
	\textbf{Why the large gain on SSv2-Full?} Our method shows a significantly stronger performance on SSv2-Full, which we attribute to two main factors. First, SSv2-Full exhibits fine-grained temporal variations absent in K100, UCF101, and HMDB51, where our task-specific matching effectively captures such detailed temporal dynamics. Second, its considerably larger base set—about 10 times that of SSv2-Small and other datasets—provides better supervision for LSN training and leads to more reliable $\alpha$-DC estimation. Together, these factors explain the larger gain on SSv2-Full.

	\subsection{Ablation Study} \label{Ablation}
	To better understand the design choices and individual contributions of our proposed TS-FSAR framework, we perform ablation studies on SSv2-Full and HMDB51.
	
	\begin{table}[htbp]
		\centering
		\setlength{\tabcolsep}{0.8mm}
		\renewcommand{\arraystretch}{1.1}
		\begin{tabular}{cccccccc}
			\toprule
			\multirow{2}{*}{LSN}
			& \multirow{2}{*}{IF-D$^{\alpha}$C}
			& \multirow{2}{*}{TSM}
			& \multirow{2}{*}{GLAC}
			& \multicolumn{2}{c}{SSv2-Full}
			& \multicolumn{2}{c}{HMDB51} \\
			& & & 
			& 1-shot & 5-shot 
			& 1-shot & 5-shot \\
			\midrule
			&       &      &     
			& 37.0    &37.0       
			& 75.9    & 75.9   \\
			\checkmark &       &       &     
			& 67.1    &77.2       
			& 77.7    & 79.6   \\
			\checkmark & \checkmark &      &     
			& 71.4    &81.7       
			& 82.1    & 84.2   \\
			\checkmark & \checkmark & \checkmark &    
			& 73.8    &82.8      
			& 83.4    & 85.6   \\
			\checkmark & \checkmark & \checkmark& \checkmark 
			& 75.1    & 83.5   
			& 85.0    & 88.9   \\
			\bottomrule
		\end{tabular}
		\caption{Ablation on key components of TS-FSAR}
		\label{ablation: key components}
	\end{table}
	
	\textbf{Ablation on Main Components} We present a comprehensive ablation study to quantify the contribution of each main component within the TS-FSAR framework, as detailed in Table~\ref{ablation: key components}. The analysis begins with the zero-shot CLIP baseline, which exhibits limited performance, achieving 37.0\% on SSv2-Full and 75.9\% on HMDB51. By introducing the LSN, we observe a marked increase in accuracy. Next, to better understand our proposed TS-DCM metric, we perform ablation by decoupling it into two components: IF-D$^{\alpha}$C and TSM. Based on the LSN, introducing IF-D$^{\alpha}$C further improves performance by 4.3\%$\sim$4.6\% across all benchmarks, demonstrating the benefit of fully capturing inter-frame dependencies. Further incorporating the TSM module yields an additional gain of 1.1\%$\sim$2.4\%, highlighting the advantage of introducing task-specific information when performing matching. Finally, the incorporation of GLAC consistently enhances performance, with especially notable improvements on HMDB51, validating our hypothesis that insufficient training of the LSN has a greater impact on static datasets. Furthermore, on SSv2-Full, removing the adapter in GLAC leads to a 3.6\% performance drop.
	
	\begin{figure}[htbp]
		\centering	
		\begin{subfigure}[b]{1.0\linewidth}
			\centering
			\includegraphics[width=0.95\textwidth]{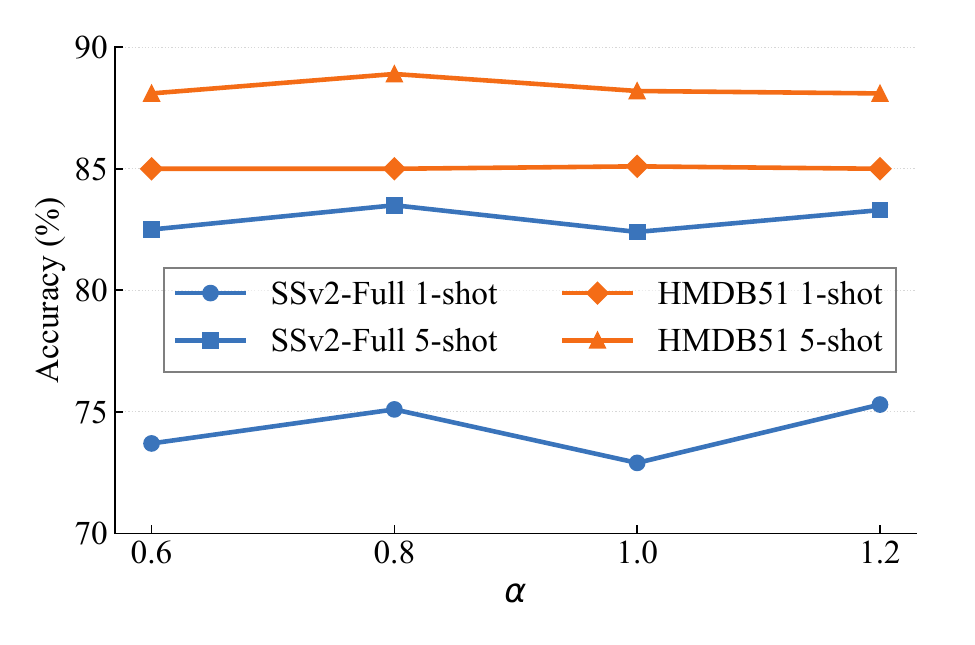}
		\end{subfigure}		
		\caption{The impact of $\alpha$ in IF-D$^{\alpha}$C} 
		\label{ablation: Impact of alpha}
	\end{figure}
	
\begin{figure}[htbp]
		\centering	
		\begin{subfigure}[b]{1.0\linewidth}
			\centering
			\includegraphics[width=0.95\textwidth]{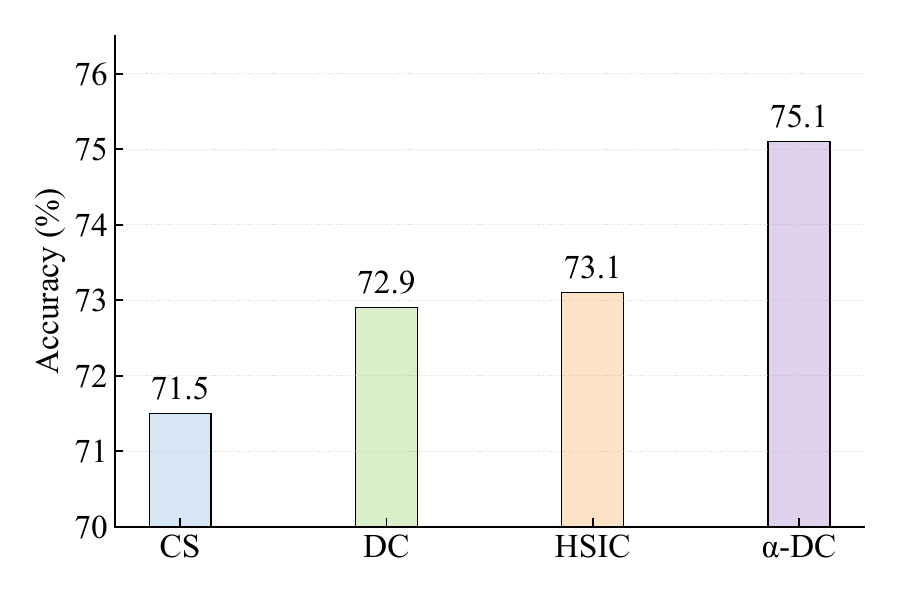}
		\end{subfigure}		
		\caption{$\alpha$-DC vs. other alternatives (1-shot on SSv2-Full)} 
		\label{figure: different_alternatives}
	\end{figure}

	\textbf{About the IF-D$^{\alpha}$C} We employ $\alpha$-DC to model inter-frame correlations, which can capture complex (nonlinear) dependencies. Similar alternatives include DC used in DeepBDC~\cite{deepbdc} and the kernel-based HSIC~\cite{hsic}. To compare them fairly, we replaced only the inter-frame similarity metric under identical settings. As shown in the Figure~\ref{figure: different_alternatives}, nonlinear measures ($\alpha$-DC, DC, HSIC) consistently outperform Cosine Similarity (CS), while $\alpha$-DC achieves the best performance owing to its more robust nonlinear modeling. As noted by~\cite{is_distance}, the coefficient $\alpha$ acts as an empirical hyperparameter that governs the trade-off between robustness and sensitivity. Hence, we further investigate the impact of $\alpha$ (Eq.~(\ref{eq: alpha})) to analyze its influence on performance. As shown in Figure~\ref{ablation: Impact of alpha}, $\alpha=1.2$ achieved the highest 1-shot accuracy on SSv2-Full, but $\alpha=0.8$ yielded the best overall results. Consequently, we selected $\alpha=0.8$ as the default throughout the paper.
	
	\begin{table}[htbp]
		\centering
		\renewcommand\arraystretch{1.1}
		\setlength\tabcolsep{3.2pt}
		\begin{tabular}{cccc}
			\toprule
			Query-Specific & Task Prototype & SSv2-Full & HMDB51 \\
			\midrule
			w/o & Average & 74.1  &84.5  \\
			\midrule
			\multirow{3}{*}{w/} & Average & 75.1 &85.0    \\
			&Concatenation &73.6  &84.6        \\
			&Cross-Attention & 74.1 &84.6        \\
			\bottomrule
		\end{tabular}
		\caption{Ablation on Query-Specific task prototype}
		\label{ablation: Query-Specific Prototype}
	\end{table}
	
	\begin{figure}[htbp]
		\centering	
		\begin{subfigure}[b]{1.0\linewidth}
			\centering
			\includegraphics[width=0.95\textwidth]{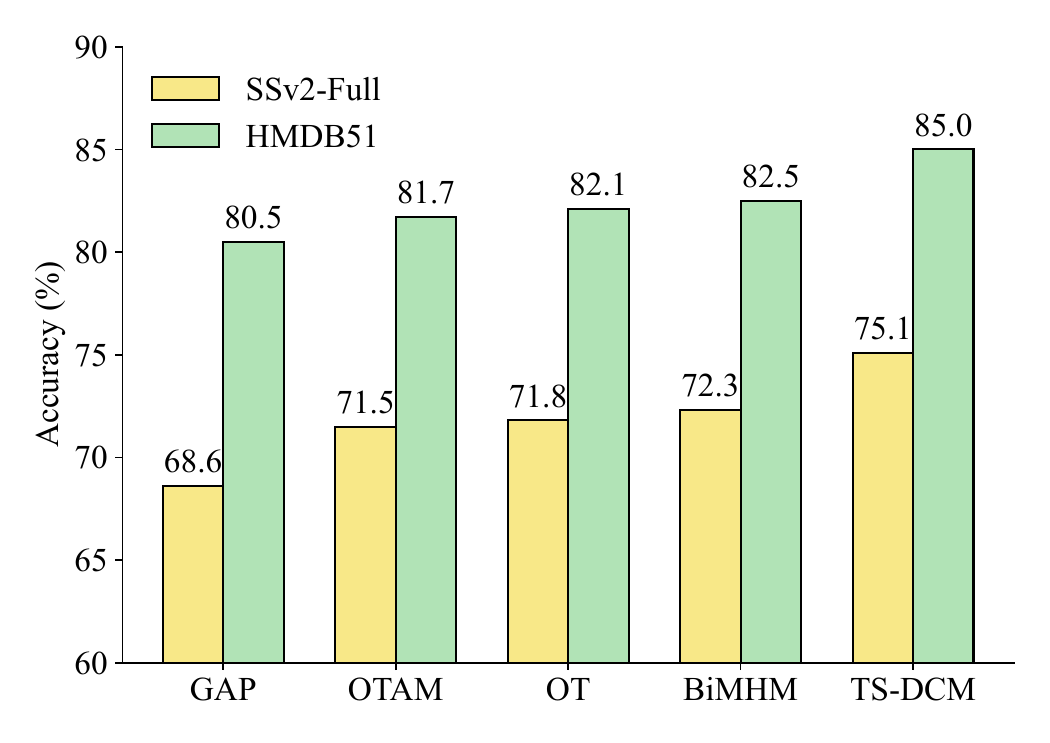}
		\end{subfigure}		
		\caption{Comparison with different metrics} 
		\label{figure: different_metrics}
	\end{figure}

	\textbf{Effect of Different Task Prototypes} To evaluate the effect of different task prototypes, we conduct a comprehensive ablation study on prototype construction strategies. Specifically, we first investigate whether incorporating the query into prototype construction is beneficial. And to build query-specific prototypes, we explore various fusion strategies between support and query class tokens, including simple averaging, concatenation along the feature dimension, and cross-attention. As shown in Table~\ref{ablation: Query-Specific Prototype}, incorporating the query into prototype construction yields a performance gain of approximately 0.5\%$\sim$1.0\% on both SSv2-Full and HMDB51. Among the query-specific strategies, simple averaging consistently outperforms concatenation and cross-attention by 0.4\%$\sim$1.5\% across both datasets.

	\textbf{Comparison of Different Metrics} To evaluate the effectiveness of the proposed TS-DCM metric, we replace it with several commonly used metrics within our framework a fair comparison. As illustrated in Figure~\ref{figure: different_metrics}, we consider Global Average Pooling (GAP), OTAM, Optimal Transport(OT), and BiMHM~\cite{hyrsm} as baselines. TS-DCM consistently achieves higher accuracy than all prior metrics on both SSv2-Full and HMDB51. In particular, compared to the second-best metric, BiMHM, TS-DCM improves performance by 2.8\% on SSv2-Full and 2.5\% on HMDB51. These results highlight the advantage of our metric in fully modeling inter-frame relationships and effectively leveraging task-specific information during matching.
	 
 	\begin{table}[htbp]
 	\centering
 	\setlength{\tabcolsep}{1.8mm}
 	\renewcommand{\arraystretch}{1.1}
 	\begin{tabular}{lcccc}
 		\toprule
 		\multirow{2}{*}{Metric}
 		& \multicolumn{2}{c}{SSv2-Full}
 		& \multicolumn{2}{c}{HMDB51} \\
 		\cmidrule(lr){2-3} \cmidrule(lr){4-5}
 		& w/o & w/  & w/o & w/ \\
 		\midrule
 		GAP   & 68.6 & 72.0 (+3.4)  & 80.5 & 81.8 (+1.3) \\
 		OTAM  & 71.5 & 72.4 (+0.9)  & 81.7 & 83.7 (+2.0) \\
 		BiMHM & 72.3 & 73.2 (+0.9)  & 82.5 & 84.5 (+2.0) \\
 		OT    & 71.8 & 73.5 (+1.7)  & 82.1 & 82.9 (+0.8) \\
 		\bottomrule
 	\end{tabular}
 	\caption{Evaluation of existing metrics with (‘w/’) and without (‘w/o’) IF-D$^{\alpha}$C.}
 	\label{tab: DC_w_others}
    \end{table}

	\begin{table}[htbp]
	\centering
	\setlength{\tabcolsep}{0.3mm}
	\renewcommand{\arraystretch}{1.1}
	\begin{tabular}{lccccc}
		\toprule
		Method & {Time} & Params & {Mem} & {SSv2-Full} & {HMDB51} \\
		\midrule
		CLIP-FSAR & 0.70 s & 89 M & $\sim$20 GB  &62.1  & 77.1 \\
		EMP-Net  & 0.45 s & 9 M & $\sim$4 GB & 63.1  & 76.8 \\
		TS-FSAR$^\dagger$  & 0.42 s & 14 M & $\sim$9 GB  & 75.1  & 85.0  \\
		TS-FSAR$^*$  & 0.32 s & 5 M & $\sim$3.6 GB  & 67.0  & 84.5  \\
		\bottomrule
	\end{tabular}
	\caption{Efficiency comparison with prior methods were performed under 5-way 1-shot setting (with 2 queries). $^*$ indicates using only a 3-layer LSN, while $^\dagger$ denotes using a 12-layer one. All evaluations were completed using a single RTX 4090 GPU.}\label{tab:training cost}
    \end{table}

	\textbf{Combine IF-D$^{\alpha}$C with existing metrics} Our proposed IF-D$^{\alpha}$C enables comprehensive modeling of inter-frame relationships. To evaluate its generalization ability, we incorporate it with existing metrics and conduct the evaluation under our setting. As shown in Table~\ref{tab: DC_w_others}, incorporating IF-D$^{\alpha}$C to model inter-frame dependencies yields performance gains of 0.8\%$\sim$3.4\% over the original metrics.

	
	\subsection{Efficiency Analysis}
	To evaluate efficiency, we report the average training and inference time per task, as well as parameter count and memory usage. As shown in Table~\ref{tab:training cost}, with a 12-layer LSN, TS-FSAR achieves the fastest runtime (0.42 s) and markedly reduces memory (9 GB) and parameters (14 M) compared to fully fine-tuned CLIP-FSAR (0.70 s, 20 GB, 89 M), while maintaining comparable cost to EMP-Net (0.45 s, 4 GB, 9 M). When the LSN depth is reduced to 3 layers, matching EMP-Net, TS-FSAR still delivers superior performance with the fastest speed, minimal memory usage, and the fewest parameters (0.32 s, 3.6 GB, 5 M).

	\section{Conclusion}

	We propose TS-FSAR, a novel few-shot action recognition framework that introduces Task-Specific Distance Correlation Matching (TS-DCM) — a new metric designed to address key limitations of previous methods, which rely on cosine similarity to model linear inter-frame dependencies and overlook task-specific cues during matching. TS-DCM uses $\alpha$-distance correlation to capture both linear and nonlinear inter-frame relationships, and employs a query-specific task prototype to enable task-specific query-support matching. To efficiently adapt CLIP, we employ a visual Ladder Side Network (LSN), whose training is guided by the adapted frozen CLIP outputs to achieve reliable correlation estimation under limited data. With these designs, TS-FSAR achieves superior performance on five standard benchmarks.
	
	\section*{Acknowledgments}
	This research was supported by the National Natural Science Foundation of China (Grant Nos. 62471083 and 61971086).
	
	\bibliography{aaai26}

@inproceedings{empnet,
  title={Efficient Few-Shot Action Recognition via Multi-level Post-reasoning},
  author={Wu, Cong and Wu, Xiao-Jun and Li, Linze and Xu, Tianyang and Feng, Zhenhua and Kittler, Josef},
  booktitle={European Conference on Computer Vision},
  pages={38--56},
  year={2024}
}

@inproceedings{cmn++,
  author       = {Linchao Zhu and Yi Yang},
  title        = {Compound Memory Networks for Few-Shot Video Classification},
  booktitle    = {Computer Vision - {ECCV} 2018 - 15th European Conference, Munich,
                  Germany, September 8-14, 2018, Proceedings, Part {VII}},
  pages        = {782--797},
  year         = {2018}
}

@inproceedings{otam,
  author       = {Kaidi Cao and
                  Jingwei Ji and
                  Zhangjie Cao and
                  Chien{-}Yi Chang and
                  Juan Carlos Niebles},
  title        = {Few-Shot Video Classification via Temporal Alignment},
  booktitle    = {2020 {IEEE/CVF} Conference on Computer Vision and Pattern Recognition,
                  {CVPR} 2020, Seattle, WA, USA, June 13-19, 2020},
  pages        = {10615--10624},
  year         = {2020}
}

@inproceedings{trx,
  author       = {Toby Perrett and
                  Alessandro Masullo and
                  Tilo Burghardt and
                  Majid Mirmehdi and
                  Dima Damen},
  title        = {Temporal-Relational CrossTransformers for Few-Shot Action Recognition},
  booktitle    = {{IEEE} Conference on Computer Vision and Pattern Recognition, {CVPR}
                  2021, virtual, June 19-25, 2021},
  pages        = {475--484},
  year         = {2021}
}

@inproceedings{strm,
  author       = {Anirudh Thatipelli and
                  Sanath Narayan and
                  Salman Khan and
                  Rao Muhammad Anwer and
                  Fahad Shahbaz Khan and
                  Bernard Ghanem},
  title        = {Spatio-temporal Relation Modeling for Few-shot Action Recognition},
  booktitle    = {{IEEE/CVF} Conference on Computer Vision and Pattern Recognition,
                  {CVPR} 2022, New Orleans, LA, USA, June 18-24, 2022},
  pages        = {19926--19935},
  year         = {2022}
}

@inproceedings{hyrsm,
  author       = {Xiang Wang and
                  Shiwei Zhang and
                  Zhiwu Qing and
                  Mingqian Tang and
                  Zhengrong Zuo and
                  Changxin Gao and
                  Rong Jin and
                  Nong Sang},
  title        = {Hybrid Relation Guided Set Matching for Few-shot Action Recognition},
  booktitle    = {{IEEE/CVF} Conference on Computer Vision and Pattern Recognition,
                  {CVPR} 2022, New Orleans, LA, USA, June 18-24, 2022},
  pages        = {19916--19925},
  year         = {2022}
}

@inproceedings{hcl,
  author       = {Sipeng Zheng and
                  Shizhe Chen and
                  Qin Jin},
  title        = {Few-Shot Action Recognition with Hierarchical Matching and Contrastive
                  Learning},
  booktitle    = {Computer Vision - {ECCV} 2022 - 17th European Conference, Tel Aviv,
                  Israel, October 23-27, 2022, Proceedings, Part {IV}},
  pages        = {297--313},
  year         = {2022}
}

@inproceedings{nguyen,
  author       = {Khoi D. Nguyen and
                  Quoc{-}Huy Tran and
                  Khoi Nguyen and
                  Binh{-}Son Hua and
                  Rang Nguyen},
  title        = {Inductive and Transductive Few-Shot Video Classification via Appearance
                  and Temporal Alignments},
  booktitle    = {Computer Vision - {ECCV} 2022 - 17th European Conference, Tel Aviv,
                  Israel, October 23-27, 2022, Proceedings, Part {XX}},
  pages        = {471--487},
  publisher    = {Springer},
  year         = {2022}
}

@inproceedings{sloshnet,
  author       = {Jiazheng Xing and
                  Mengmeng Wang and
                  Yong Liu and
                  Boyu Mu},
  title        = {Revisiting the Spatial and Temporal Modeling for Few-Shot Action Recognition},
  booktitle    = {Thirty-Seventh {AAAI} Conference on Artificial Intelligence, {AAAI}
                  2023, Thirty-Fifth Conference on Innovative Applications of Artificial
                  Intelligence, {IAAI} 2023, Thirteenth Symposium on Educational Advances
                  in Artificial Intelligence, {EAAI} 2023, Washington, DC, USA, February
                  7-14, 2023},
  pages        = {3001--3009},
  year         = {2023}
}

@inproceedings{gghm,
  author       = {Jiazheng Xing and
                  Mengmeng Wang and
                  Yudi Ruan and
                  Bofan Chen and
                  Yaowei Guo and
                  Boyu Mu and
                  Guang Dai and
                  Jingdong Wang and
                  Yong Liu},
  title        = {Boosting Few-shot Action Recognition with Graph-guided Hybrid Matching},
  booktitle    = {{IEEE/CVF} International Conference on Computer Vision, {ICCV} 2023,
                  Paris, France, October 1-6, 2023},
  pages        = {1740--1750},
  year         = {2023}
}

@article{clipfsar,
  author       = {Xiang Wang and
                  Shiwei Zhang and
                  Jun Cen and
                  Changxin Gao and
                  Yingya Zhang and
                  Deli Zhao and
                  Nong Sang},
  title        = {CLIP-guided Prototype Modulating for Few-shot Action Recognition},
  journal      = {Int. J. Comput. Vis.},
  pages        = {1899--1912},
  year         = {2024}
}

@inproceedings{ds2tadapter,
  title={D $^2$ ST-Adapter: Disentangled-and-Deformable Spatio-Temporal Adapter for Few-shot Action Recognition},
  author={Pei, Wenjie and Tan, Qizhong and Lu, Guangming and Tian, Jiandong},
  booktitle={{IEEE/CVF} International Conference on Computer Vision},
  year={2025}
}

@inproceedings{tsam,
  author       = {Bozheng Li and
                  Mushui Liu and
                  Gaoang Wang and
                  Yunlong Yu},
  title        = {Frame Order Matters: {A} Temporal Sequence-Aware Model for Few-Shot
                  Action Recognition},
  booktitle    = {AAAI-25, Sponsored by the Association for the Advancement of Artificial
                  Intelligence, February 25 - March 4, 2025, Philadelphia, PA, {USA}},
  pages        = {18218--18226},
  year         = {2025}
}

@inproceedings{lst,
  author       = {Yi{-}Lin Sung and
                  Jaemin Cho and
                  Mohit Bansal},
  title        = {{LST:} Ladder Side-Tuning for Parameter and Memory Efficient Transfer
                  Learning},
  booktitle    = {Advances in Neural Information Processing Systems 35: Annual Conference
                  on Neural Information Processing Systems 2022, NeurIPS 2022, New Orleans,
                  LA, USA, November 28 - December 9, 2022},
  year         = {2022}
}

@inproceedings{ssv2,
  author       = {Raghav Goyal and
                  Samira Ebrahimi Kahou and
                  Vincent Michalski and
                  Joanna Materzynska and
                  Susanne Westphal and
                  Heuna Kim and
                  Valentin Haenel and
                  Ingo Fr{\"{u}}nd and
                  Peter Yianilos and
                  Moritz Mueller{-}Freitag and
                  Florian Hoppe and
                  Christian Thurau and
                  Ingo Bax and
                  Roland Memisevic},
  title        = {The "Something Something" Video Database for Learning and Evaluating
                  Visual Common Sense},
  booktitle    = {{IEEE} International Conference on Computer Vision, {ICCV} 2017, Venice,
                  Italy, October 22-29, 2017},
  pages        = {5843--5851},
  year         = {2017}
}

@inproceedings{k400,
  author       = {Jo{\~{a}}o Carreira and
                  Andrew Zisserman},
  title        = {Quo Vadis, Action Recognition? {A} New Model and the Kinetics Dataset},
  booktitle    = {2017 {IEEE} Conference on Computer Vision and Pattern Recognition,
                  {CVPR} 2017, Honolulu, HI, USA, July 21-26, 2017},
  pages        = {4724--4733},
  year         = {2017}
}

@inproceedings{hmdb51,
  author       = {Hildegard Kuehne and
                  Hueihan Jhuang and
                  Est{\'{\i}}baliz Garrote and
                  Tomaso A. Poggio and
                  Thomas Serre},
  title        = {{HMDB:} {A} large video database for human motion recognition},
  booktitle    = {{IEEE} International Conference on Computer Vision, {ICCV} 2011, Barcelona,
                  Spain, November 6-13, 2011},
  pages        = {2556--2563},
  year         = {2011}
}

@article{ucf101,
  title={UCF101: A dataset of 101 human actions classes from videos in the wild},
  author={Soomro, Khurram and Zamir, Amir Roshan and Shah, Mubarak},
  journal={arXiv preprint arXiv:1212.0402},
  year={2012}
}

@inproceedings{mtfan,
  author       = {Jiamin Wu and
                  Tianzhu Zhang and
                  Zhe Zhang and
                  Feng Wu and
                  Yongdong Zhang},
  title        = {Motion-modulated Temporal Fragment Alignment Network For Few-Shot
                  Action Recognition},
  booktitle    = {{IEEE/CVF} Conference on Computer Vision and Pattern Recognition,
                  {CVPR} 2022, New Orleans, LA, USA, June 18-24, 2022},
  pages        = {9141--9150},
  year         = {2022}
}

@article{bdc,
  author  = {Sz\'ekely, G\'abor J. and Rizzo, Maria L},
  title   = {Brownian distance covariance},
  journal = {Annals of Statistics},
  year    = {2009},
  volume  = {3},
  pages   = {1236--1265}
}

@inproceedings{sidetuning,
  title={Side-tuning: a baseline for network adaptation via additive side networks},
  author={Zhang, Jeffrey O and Sax, Alexander and Zamir, Amir and Guibas, Leonidas and Malik, Jitendra},
  booktitle={Computer Vision--ECCV 2020: 16th European Conference, Glasgow, UK, August 23--28, 2020, Proceedings, Part III 16},
  pages={698--714},
  year={2020}
}

@inproceedings{evl,
  title={Frozen clip models are efficient video learners},
  author={Lin, Ziyi and Geng, Shijie and Zhang, Renrui and Gao, Peng and De Melo, Gerard and Wang, Xiaogang and Dai, Jifeng and Qiao, Yu and Li, Hongsheng},
  booktitle={European Conference on Computer Vision},
  pages={388--404},
  year={2022}
}

@inproceedings{stan,
  title={Revisiting temporal modeling for clip-based image-to-video knowledge transferring},
  author={Liu, Ruyang and Huang, Jingjia and Li, Ge and Feng, Jiashi and Wu, Xinglong and Li, Thomas H},
  booktitle={Proceedings of the IEEE/CVF Conference on Computer Vision and Pattern Recognition},
  pages={6555--6564},
  year={2023}
}

@article{mafsar,
title = {MA-FSAR: Multimodal Adaptation of CLIP for few-shot action recognition},
journal = {Pattern Recognition},
pages = {111902},
year = {2025},
author = {Jiazheng Xing and Jian Zhao and Chao Xu and Mengmeng Wang and Guang Dai and Yong Liu and Jingdong Wang and Xuelong Li}
}

@inproceedings{trajectory,
  title={Trajectory-aligned Space-time Tokens for Few-shot Action Recognition},
  author={Kumar, Pulkit and Padmanabhan, Namitha and Luo, Luke and Rambhatla, Sai Saketh and Shrivastava, Abhinav},
  booktitle={European Conference on Computer Vision},
  pages={474--493},
  year={2024}
}

@inproceedings{team,
  title={Temporal Alignment-Free Video Matching for Few-shot Action Recognition},
  author={Lee, SuBeen and Moon, WonJun and Seong, Hyun Seok and Heo, Jae-Pil},
  booktitle={Proceedings of the Computer Vision and Pattern Recognition Conference},
  pages={5412--5421},
  year={2025}
}

@inproceedings{task_adapter,
  title={Task-Adapter: Task-specific Adaptation of Image Models for Few-shot Action Recognition},
  author={Cao, Congqi and Zhang, Yueran and Yu, Yating and Lv, Qinyi and Min, Lingtong and Zhang, Yanning},
  booktitle={Proceedings of the 32nd ACM International Conference on Multimedia},
  pages={9038--9047},
  year={2024}
}

@article{is_distance,
  title={Is Distance Correlation Robust?”},
  author={Leyder, Sarah and Raymaekers, Jakob and Rousseeuw, Peter J},
  journal={arXiv preprint arXiv:2403.03722},
  volume={122},
  year={2024}
}

@inproceedings{deepbdc,
  author       = {Jiangtao Xie and
                  Fei Long and
                  Jiaming Lv and
                  Qilong Wang and
                  Peihua Li},
  title        = {Joint Distribution Matters: Deep Brownian Distance Covariance for
                  Few-Shot Classification},
  booktitle    = {{IEEE/CVF} Conference on Computer Vision and Pattern Recognition,
                  {CVPR} 2022, New Orleans, LA, USA, June 18-24, 2022},
  pages        = {7962--7971},
  year         = {2022}
}

@inproceedings{vivit,
  title={Vivit: A video vision transformer},
  author={Arnab, Anurag and Dehghani, Mostafa and Heigold, Georg and Sun, Chen and Lu{\v{c}}i{\'c}, Mario and Schmid, Cordelia},
  booktitle={Proceedings of the IEEE/CVF international conference on computer vision},
  pages={6836--6846},
  year={2021}
}

@inproceedings{tsn,
  title={Temporal segment networks: Towards good practices for deep action recognition},
  author={Wang, Limin and Xiong, Yuanjun and Wang, Zhe and Qiao, Yu and Lin, Dahua and Tang, Xiaoou and Van Gool, Luc},
  booktitle={European conference on computer vision},
  pages={20--36},
  year={2016}
}

@inproceedings{nonlocal,
  title={Non-local neural networks},
  author={Wang, Xiaolong and Girshick, Ross and Gupta, Abhinav and He, Kaiming},
  booktitle={Proceedings of the IEEE conference on computer vision and pattern recognition},
  pages={7794--7803},
  year={2018}
}

@inproceedings{tsm,
  title={Tsm: Temporal shift module for efficient video understanding},
  author={Lin, Ji and Gan, Chuang and Han, Song},
  booktitle={Proceedings of the IEEE/CVF international conference on computer vision},
  pages={7083--7093},
  year={2019}
}

@inproceedings{i3d,
  title={Quo vadis, action recognition? a new model and the kinetics dataset},
  author={Carreira, Joao and Zisserman, Andrew},
  booktitle={proceedings of the IEEE Conference on Computer Vision and Pattern Recognition},
  pages={6299--6308},
  year={2017}
}

@inproceedings{trn,
  title={Temporal relational reasoning in videos},
  author={Zhou, Bolei and Andonian, Alex and Oliva, Aude and Torralba, Antonio},
  booktitle={Proceedings of the European conference on computer vision (ECCV)},
  pages={803--818},
  year={2018}
}

@inproceedings{clip,
  author       = {Alec Radford and
                  Jong Wook Kim and
                  Chris Hallacy and
                  Aditya Ramesh and
                  Gabriel Goh and
                  Sandhini Agarwal and
                  Girish Sastry and
                  Amanda Askell and
                  Pamela Mishkin and
                  Jack Clark and
                  Gretchen Krueger and
                  Ilya Sutskever},
  title        = {Learning Transferable Visual Models From Natural Language Supervision},
  booktitle    = {Proceedings of the 38th International Conference on Machine Learning,
                  {ICML} 2021, 18-24 July 2021, Virtual Event},
  pages        = {8748--8763},
  year         = {2021}
}

@inproceedings{correlation,
	title={Correlation Coefficients and Semantic Textual Similarity},
	author={Zhelezniak, Vitalii and Savkov, Aleksandar and Shen, April and Hammerla, Nils Y.},
	booktitle={Conference of the North American Chapter of the Association for Computational Linguistics: Human Language Technologies },
	pages={951--962},
	year={2019},
}

@inproceedings{dual_path,
  title={Dual-path adaptation from image to video transformers},
  author={Park, Jungin and Lee, Jiyoung and Sohn, Kwanghoon},
  booktitle={Proceedings of the IEEE/CVF Conference on Computer Vision and Pattern Recognition},
  pages={2203--2213},
  year={2023}
}

@article{mvp,
  title={MVP-shot: Multi-velocity progressive-alignment framework for few-shot action recognition},
  author={Qu, Hongyu and Yan, Rui and Shu, Xiangbo and Gao, Hailiang and Huang, Peng and Xie, Guo-Sen},
  journal={IEEE Transactions on Multimedia},
  year={2025},
  publisher={IEEE}
}

@inproceedings{riva,
  title={Rethinking image-to-video adaptation: An object-centric perspective},
  author={Qian, Rui and Ding, Shuangrui and Lin, Dahua},
  booktitle={European Conference on Computer Vision},
  pages={329--348},
  year={2024},
  organization={Springer}
}

@inproceedings{resnet,
  title={Deep residual learning for image recognition},
  author={He, Kaiming and Zhang, Xiangyu and Ren, Shaoqing and Sun, Jian},
  booktitle={Proceedings of the IEEE conference on computer vision and pattern recognition},
  pages={770--778},
  year={2016}
}

@inproceedings{vit,
  author       = {Alexey Dosovitskiy and
                  Lucas Beyer and
                  Alexander Kolesnikov and
                  Dirk Weissenborn and
                  Xiaohua Zhai and
                  Thomas Unterthiner and
                  Mostafa Dehghani and
                  Matthias Minderer and
                  Georg Heigold and
                  Sylvain Gelly and
                  Jakob Uszkoreit and
                  Neil Houlsby},
  title        = {An Image is Worth 16x16 Words: Transformers for Image Recognition
                  at Scale},
  booktitle    = {9th International Conference on Learning Representations, {ICLR} 2021,
                  Virtual Event, Austria, May 3-7, 2021},
  publisher    = {OpenReview.net},
  year         = {2021}
}

@inproceedings{ta2n,
  title={Ta2n: Two-stage action alignment network for few-shot action recognition},
  author={Li, Shuyuan and Liu, Huabin and Qian, Rui and Li, Yuxi and See, John and Fei, Mengjuan and Yu, Xiaoyuan and Lin, Weiyao},
  booktitle={Proceedings of the AAAI conference on artificial intelligence},
  pages={1404--1411},
  year={2022}
}

@inproceedings{depth,
  title={Depth guided adaptive meta-fusion network for few-shot video recognition},
  author={Fu, Yuqian and Zhang, Li and Wang, Junke and Fu, Yanwei and Jiang, Yu-Gang},
  booktitle={Proceedings of the 28th ACM international conference on multimedia},
  pages={1142--1151},
  year={2020}
}

@inproceedings{molo,
  title={Molo: Motion-augmented long-short contrastive learning for few-shot action recognition},
  author={Wang, Xiang and Zhang, Shiwei and Qing, Zhiwu and Gao, Changxin and Zhang, Yingya and Zhao, Deli and Sang, Nong},
  booktitle={Proceedings of the IEEE/CVF conference on computer vision and pattern recognition},
  pages={18011--18021},
  year={2023}
}

@inproceedings{hsic,
  title={Measuring statistical dependence with Hilbert-Schmidt norms},
  author={Gretton, Arthur and Bousquet, Olivier and Smola, Alex and Sch{\"o}lkopf, Bernhard},
  booktitle={International conference on algorithmic learning theory},
  pages={63--77},
  year={2005},
  organization={Springer}
}
	
\end{document}